\documentclass[10pt,twocolumn,letterpaper]{article}
\usepackage{cvpr}
\usepackage{mathtools,microtype,multirow,array}
\definecolor{favrteal}{HTML}{007F79}
\definecolor{favrblue}{HTML}{41577C}
\usepackage[pagebackref,breaklinks,colorlinks,citecolor=favrblue,linkcolor=favrblue,urlcolor=favrblue]{hyperref}

\hypersetup{pdftitle={FAVR: Flow-Aware Visual Residual Alignment for Vision-Language-Action Policies},pdfauthor={Qixuan Wu, Xiaole Yang, Yuguang Yang, Canyu Chen, Linlin Yang, Haixia Pan, Guodong Guo, Baochang Zhang}}
\makeatletter
\patchcmd{\@maketitle}{\vskip .375in}{\vskip .16in}{}{}
\patchcmd{\@maketitle}{\vspace*{24pt}}{\vspace*{13pt}}{}{}
\makeatother
\usepackage{placeins}
\def\confName{CVPR}
\def\confYear{2027}
\title{FAVR: Flow-Aware Visual Residual Alignment\\for Vision-Language-Action Policies}
\author{%
\begin{minipage}{0.99\textwidth}\centering
{\large Qixuan Wu$^{1,*}$\quad Xiaole Yang$^{2,*}$\quad Yuguang Yang$^{3,\dagger}$\quad Canyu Chen$^4$}\\[2pt]
{\large Linlin Yang$^5$\quad Haixia Pan$^2$\quad Guodong Guo$^6$\quad Baochang Zhang$^{1,\dagger}$}\\[5pt]
{\fontsize{8}{9.25}\selectfont
$^1$School of Artificial Intelligence, Beihang University, Beijing, China\\
$^2$School of Software, Beihang University, Beijing, China\\
$^3$School of Electronic and Information Engineering, Beihang University, Beijing, China\\
$^4$National College for Excellent Engineers, Beihang University, Beijing, China\\
$^5$Communication University of China, Beijing, China\\
$^6$Lane Department of Computer Science and Electrical Engineering, West Virginia University, Morgantown, USA\\
}
\vspace{0.20em}
{\small $^*$Equal contribution.\qquad $^\dagger$Corresponding authors.}
\end{minipage}}
\begin{document}
{\setlength{\tabcolsep}{2pt}\maketitle}
\begin{abstract}
Pretrained video representations provide rich temporal information for robotic manipulation, yet effectively integrating them into existing vision-language-action (VLA) policies remains challenging. We present Flow-Aware Visual Residual Alignment (FAVR), a lightweight framework for flow-aware visual conditioning of frozen VLA policies. FAVR maintains a compact temporal visual state through linear Test-Time Training (TTT) and retrieves action-conditioned visual features throughout flow-based action generation. By exploiting the affine structure of the frozen action head, we explicitly characterize how injected visual features influence predicted action velocities and formulate flow-aware channel selection as a box-constrained optimization problem. We develop a FISTA-based solver for joint gate optimization and derive an efficient closed-form approximation through diagonalization and rescaling. With only 1.74--2.19M trainable parameters and 5,000 adaptation steps, FAVR improves pretrained VLA policies across benchmarks. On RoboTwin 2.0, FAVR improves LingBot-VLA from 87.12\% to 90.04\% under clean conditions and from 86.88\% to 89.38\% under randomized conditions. A controlled ablation on five RoboTwin tasks further shows that flow-aware gating increases average success from 72.0\% with a gate shared across flow steps to 82.0\%, demonstrating the benefit of adapting visual conditioning throughout action generation.
\end{abstract}

\section{Introduction}
\label{sec:intro}

\begin{figure}[t]
    \centering
    \includegraphics[width=0.90\linewidth]{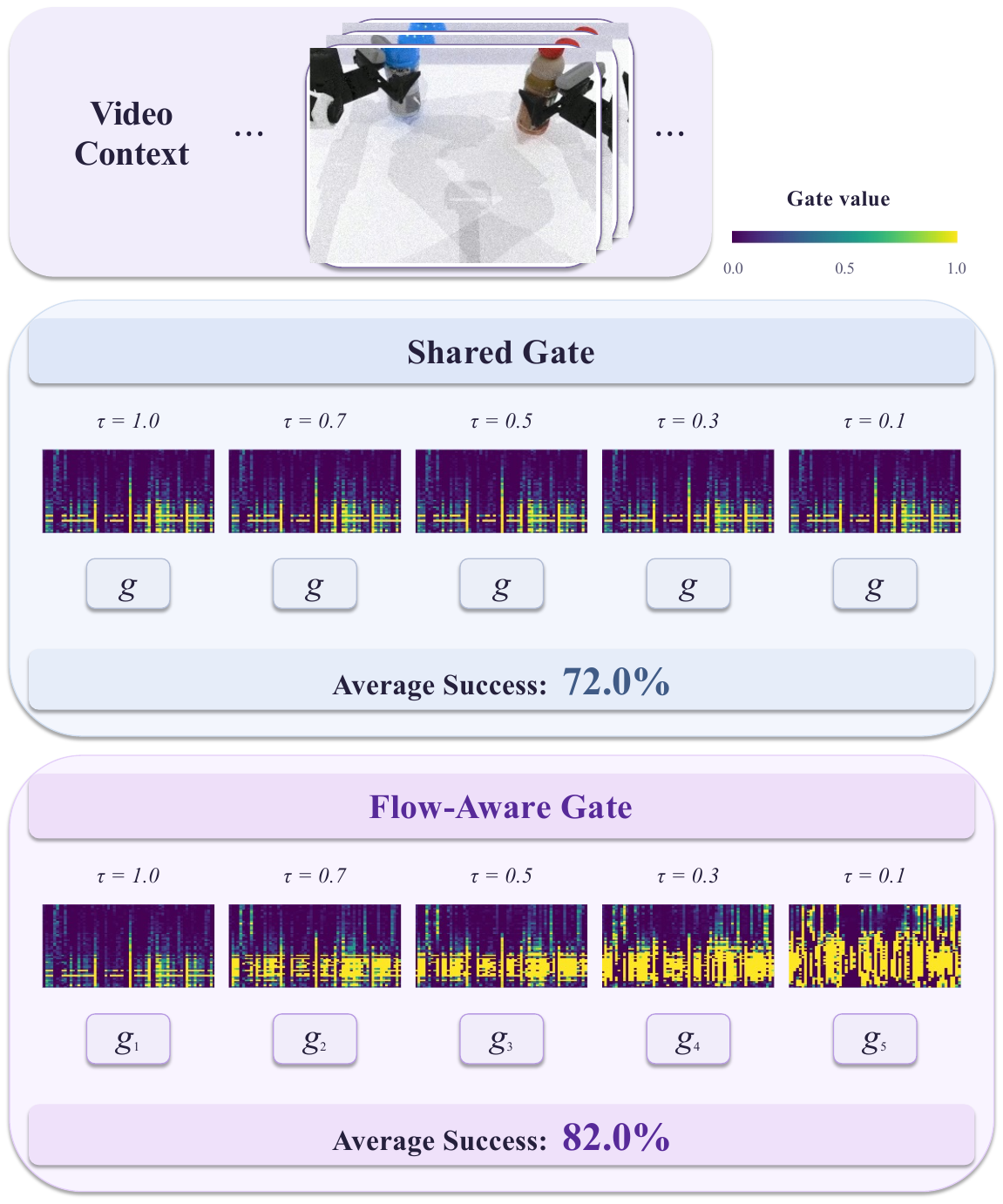}
    \caption{Shared versus flow-aware gating under the same video context. The heatmaps show recorded channel weights at representative flow steps. On five RoboTwin 2.0 tasks, average success increases from 72.0\% with a shared gate to 82.0\%.}
    \label{fig:flow_aware_motivation}
\end{figure}

Vision-language-action (VLA) models adapt pretrained vision-language representations to robotic control. Flow-based policies such as $\pi_0$ and $\pi_{0.5}$ generate continuous action chunks conditioned on visual observations and language instructions~\cite{pi0,pi05}. Video pretraining provides a complementary source of visual knowledge by learning representations of scene structure, object interactions, and temporal evolution~\cite{vjepa,vjepa2,vjepa21}. Recent studies demonstrate the benefits of incorporating video representations into VLA policies~\cite{jepavla,pvi}, raising the question of how to effectively use them within existing action policies.

Incorporating video representations requires more than aggregating temporal information and retrieving action-relevant features: the retrieved features must also benefit action generation. For flow-based policies, action chunks are constructed through successive velocity evaluations. Although the visual context remains fixed within an action chunk, the noisy action representation and corresponding policy features evolve across flow steps, so the appropriate visual contribution may also change. While state-dependent retrieval can implicitly learn this contribution, it does not explicitly account for the induced corrections in the action-velocity space. This motivates explicit velocity-residual alignment across flow steps.

We introduce \emph{Flow-Aware Visual Residual Alignment} (FAVR), a lightweight framework that integrates pretrained video representations into frozen VLA policies. FAVR employs linear Test-Time Training (TTT)~\cite{ttt} to aggregate features from a frozen video encoder into a compact temporal visual state, queried by the current action features at each flow step. Learned projections adapt state construction and retrieval to the control task. A lightweight predictor estimates the target velocity residual, while the frozen affine action head provides an exact mapping from gated visual features to velocity corrections. We formulate flow-aware channel selection as a box-constrained velocity-residual alignment problem and employ a FISTA-based solver to jointly optimize channel weights while accounting for their interactions. The gated visual features are injected through a residual connection immediately before the action head. Both pretrained backbones remain frozen, with only the auxiliary visual interface and residual predictor trained.

Joint gate optimization, however, requires iterative computation at each flow step. To reduce this overhead, we derive an approximate closed-form gate through diagonalization and analytic rescaling, simplifying channel interactions and calibrating the resulting correction. A coupling-dependent error bound characterizes the effect of the omitted interactions. The approximation reduces computational cost at the expense of a modest portion of the joint solver's performance gains.
As illustrated in Fig.~\ref{fig:flow_aware_motivation}, the flow-aware gating employed by FAVR adapts channel weights across flow steps, unlike a shared gate. On five RoboTwin 2.0 tasks, this design improves average success from 72.0\% to 82.0\%.

We evaluate FAVR on LIBERO, LIBERO-Plus, RoboTwin 2.0, and real-world dual-arm manipulation tasks. With only 1.74--2.19M additional trainable parameters and 5,000 adaptation steps in simulation, FAVR achieves 98.6\% and 85.05\% success on LIBERO and LIBERO-Plus, respectively, using $\pi_{0.5}$. On RoboTwin 2.0, it improves LingBot-VLA from 87.12\% to 90.04\% under clean conditions and from 86.88\% to 89.38\% under randomized conditions. Real-world experiments further show improvements in block pick-and-place and paper-cup grasping over continued fine-tuning.

Our main contributions are summarized as follows:
\begin{itemize}
    \item We introduce FAVR, a lightweight framework that integrates temporal video representations into frozen VLA policies through action-conditioned retrieval and flow-aware visual conditioning.

    \item We formulate visual channel selection as a box-constrained velocity-residual alignment problem, with a FISTA-based joint solver and an approximate closed-form solution supported by a coupling-dependent error bound.

    \item We validate FAVR across three manipulation benchmarks and real-world experiments, with controlled ablations demonstrating the benefits of flow-aware gating and velocity-residual alignment.
\end{itemize}

\begin{figure*}[t]
    \centering
    \includegraphics[width=\textwidth]{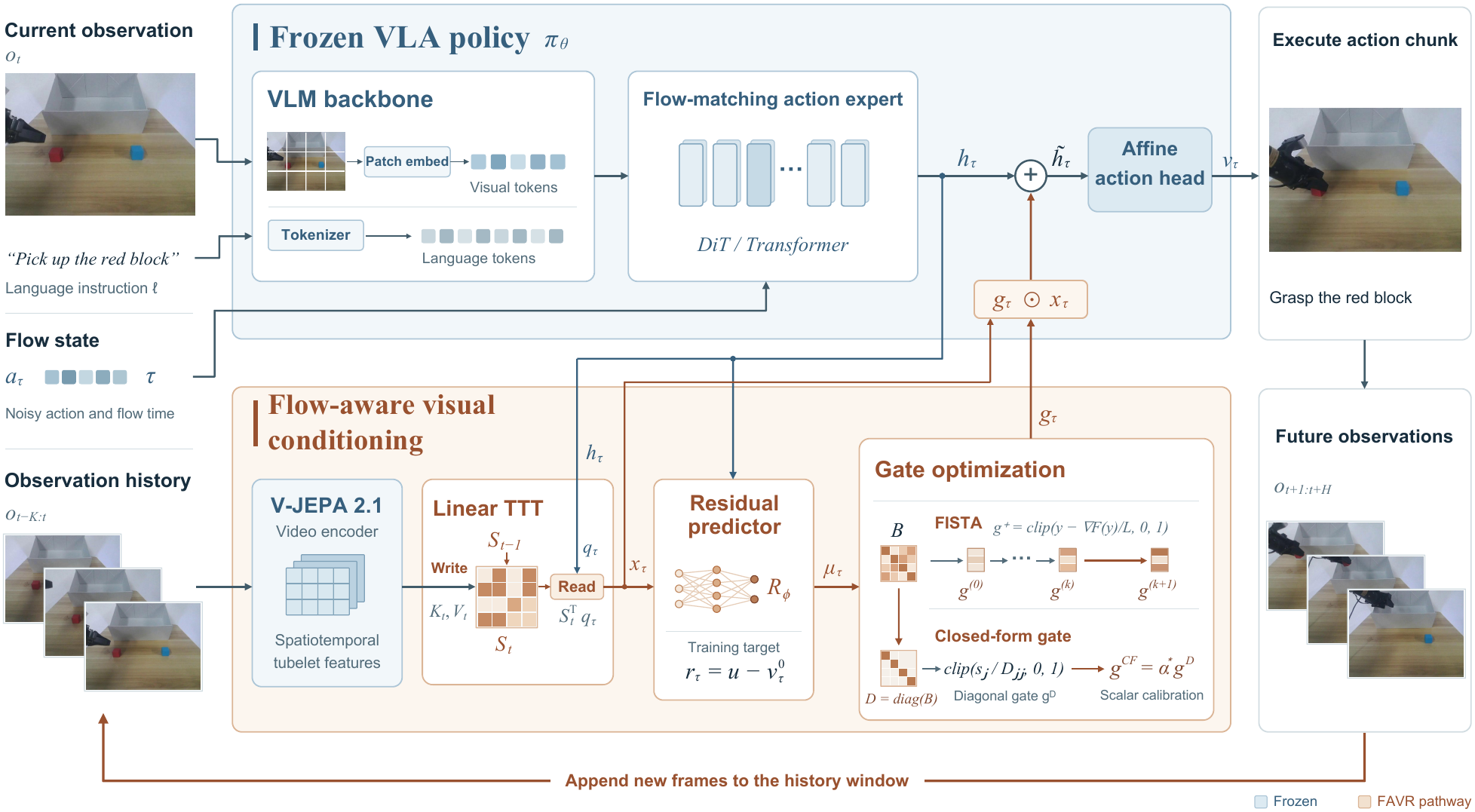}
    \caption{Overview of FAVR. Observed video features update a linear TTT state queried by action features at each flow step. A predicted velocity residual guides FISTA-based joint or approximate closed-form gating before the frozen affine action head. Executed actions provide subsequent observations for state updates.}
    \label{fig:favr_framework}
\end{figure*}

\section{Related Work}
\label{sec:related_work}

\noindent\textbf{{Vision-Language-Action Models.}}
Vision-language-action (VLA) models transfer pretrained visual and language representations to robotic manipulation, as exemplified by OpenVLA~\cite{openvla}. OpenVLA-OFT improves action generation through parallel decoding, action chunking, and continuous regression~\cite{oft}. Flow-based policies such as $\pi_0$~\cite{pi0}, $\pi_{0.5}$~\cite{pi05}, and LingBot-VLA~\cite{lingbot} combine pretrained vision-language backbones with action experts trained using flow matching~\cite{flowmatching}, generating continuous action chunks through successive velocity evaluations.

\noindent\textbf{Video Representations for Robot Control.}
Video pretraining captures scene structure and temporal dynamics, as demonstrated by the V-JEPA series~\cite{vjepa,vjepa2,vjepa21}. WorldVLA and DreamVLA incorporate visual prediction into action generation through autoregressive modeling and diffusion conditioning, respectively~\cite{worldvla,dreamvla}. JEPA-VLA integrates frozen V-JEPA~2 features through input fusion or gated cross-attention~\cite{jepavla}, while PVI injects auxiliary visual features through a trainable action expert conditioned on the current noisy action~\cite{pvi}. Both studies show that video-pretrained representations can outperform image-pretrained features in robotic manipulation~\cite{jepavla,pvi}.

\noindent\textbf{Test-Time Training for Robot Policies.}
Test-time training (TTT) maintains recurrent states through self-supervised fast-weight updates~\cite{ttt}. Recent studies have extended TTT to robotic policies, enabling the aggregation of historical observations and action-generation features to support temporally informed control~\cite{robottt,recastvla,wamttt,tttrm,t2mem,ttt_in_vla_2026}. Unlike methods that incorporate fast-weight updates within pretrained policy components, FAVR employs linear TTT as an external temporal visual interface, aggregating features from a frozen video encoder and retrieving them using current action features at each flow step.

\noindent\textbf{Time-Dependent Conditioning in Diffusion and Flow Models.}
Diffusion Policy conditions action denoising on noisy actions and diffusion timesteps~\cite{diffusionpolicy}. Beyond classifier-free guidance (CFG)~\cite{cfg}, prior work explores noise-dependent guidance intervals~\cite{limited_guidance}, condition annealing~\cite{cads}, and dynamically adjusted guidance scales~\cite{dynamic_cfg,adaptive_cfg}. Related studies analyze guidance in probability-flow dynamics and flow matching~\cite{analytic_cfg,flow_guidance}. These works motivate time-dependent conditioning~\cite{date2025}, while FAVR explicitly optimizes channel-wise visual contributions according to their induced velocity corrections through a frozen VLA action head.

\section{Method}
\label{sec:method}

FAVR augments a frozen flow-based VLA policy with temporal visual conditioning and flow-aware residual alignment (Fig.~\ref{fig:favr_framework}). At planning step $t$, the frozen policy $\pi_\theta$ conditions action generation on the current observation $o_t$, proprioception, and language instruction $\ell$. Following the rectified-flow formulation~\cite{flowmatching,liu2023rectifiedflow,pi0,pi05}, a normalized demonstration action chunk $a\in\mathbb{R}^{H\times m}$ and Gaussian noise $\epsilon$ of the same shape define
\begin{equation}
a_\tau=(1-\tau)a+\tau\epsilon,
\qquad u=\epsilon-a,
\label{eq:favr_flow}
\end{equation}

where $H$ is the action-chunk length, $m$ the action dimension, $\tau\in[0,1]$ the flow time, and $u$ the target velocity. Conditioned on $a_\tau$ and $\tau$, the frozen action expert produces a final normalized feature $h_\tau\in\mathbb{R}^d$, followed by an affine action head $v_\tau^0=Wh_\tau+b$, where $W\in\mathbb{R}^{m\times d}$ and $b\in\mathbb{R}^m$ are frozen.
For simplicity, feature- and velocity-space equations refer to one action position within the jointly generated chunk, with its index omitted; $u$ below denotes the corresponding target velocity. FAVR retrieves temporal visual information using $h_\tau$. Only the visual interface and residual predictor are trained.

\subsection{Temporal Visual State Construction and Retrieval}
\label{sec:favr_interface}

For each observed video window, a frozen V-JEPA~2.1 encoder~\cite{vjepa21} extracts spatiotemporal visual tokens $Z$. 
Learned mappings $f_K$ and $f_V$ produce keys and values, denoted $K_t$ and $V_t$ for the newly observed video features in Fig.~\ref{fig:favr_framework}. For a token block with $K,V\in\mathbb{R}^{n\times p}$, linear TTT~\cite{ttt} maintains a state $S\in\mathbb{R}^{p\times p}$ by fitting the key-to-value mapping. One gradient step on $\|KS-V\|_F^2/(2n)$ gives
\begin{equation}
S\leftarrow S-\frac{\eta}{n}K^\top(KS-V),
\label{eq:favr_write}
\end{equation}
where $\eta=\operatorname{sigmoid}(\eta_{\mathrm{raw}})$ and the projections are learned during adaptation. Each environment maintains an independent state initialized from learned $S_0$ at episode boundaries. At planning step $t$, the working state $S$ starts from $S_{t-1}$ and is updated with newly observed token blocks through Eq.~\eqref{eq:favr_write}. The resulting state $S_t$ remains fixed during action generation and is updated again when new observations arrive.

At each flow evaluation, the current action feature $h_\tau$ queries $S_t$. Treating vectors as columns, the query and readout are
\begin{equation}
\begin{aligned}
q_\tau
&=\operatorname{norm}\!\left(
P_Q^\top\operatorname{LN}(h_\tau)
\right),\\
\widetilde{x}_\tau
&=P_O^\top S_t^\top q_\tau+b_O,
\end{aligned}
\label{eq:favr_read}
\end{equation}
where $\operatorname{LN}$ denotes layer normalization, $\operatorname{norm}$ denotes $\ell_2$ normalization, $P_Q\in\mathbb{R}^{d\times p}$ and $P_O\in\mathbb{R}^{p\times d}$ are learned projections, and $b_O$ is a learned bias. We constrain the readout magnitude relative to $h_\tau$:
\begin{equation}
x_\tau=\widetilde{x}_\tau\,
\operatorname{clip}_{[0,1]}\!\left(
\frac{\rho\,\operatorname{sg}(\|h_\tau\|_2)}
{\|\widetilde{x}_\tau\|_2+\varepsilon_{\mathrm{num}}}
\right),
\label{eq:favr_normcap}
\end{equation}
where $\operatorname{sg}$ stops gradients, $\rho>0$ controls the relative magnitude, and $\varepsilon_{\mathrm{num}}>0$ ensures numerical stability. The retrieved feature $x_\tau$ is subsequently gated and injected before the frozen action head.

\subsection{Flow-Aware Visual Residual Alignment}
\label{sec:favr_alignment}

Although $S_t$ remains fixed during action generation, the noisy action $a_\tau$ and corresponding policy features evolve across flow steps. Consequently, both the visual readout $x_\tau$ and base velocity $v_\tau^0$ change with $\tau$, motivating flow-dependent visual corrections.

During training, the target velocity $u$ defines the residual $r_\tau=u-v_\tau^0$. Since $u$ is unavailable at inference, a lightweight predictor $R_\phi$ estimates $\mu_\tau=R_\phi(h_\tau,x_\tau,Wx_\tau,e(\tau))$ from the current action feature, visual readout, its velocity-space projection, and flow-time embedding. The predictor is trained against the detached residual using valid-coordinate MSE, with its inputs detached during this update. The predicted residual guides gating during both training and inference.

For $g_\tau\in[0,1]^d$, the gated visual readout produces the corrected feature $\widetilde h_\tau=h_\tau+g_\tau\odot x_\tau$. The frozen affine head then gives
\begin{equation}
\begin{aligned}
v_\tau
&=W\widetilde h_\tau+b\\
&=v_\tau^0+Q_\tau g_\tau,
\qquad Q_\tau=W\operatorname{diag}(x_\tau).
\end{aligned}
\label{eq:favr_head}
\end{equation}
Each column $x_{\tau,j}W_{:j}$ describes one visual channel's contribution to the predicted velocity. We align the induced correction with the predicted residual by solving
\begin{equation}
g_\tau^\star\in
\underset{g\in[0,1]^d}{\arg\min}
\;F_\tau(g),
\qquad
F_\tau(g)=\frac12\|Q_\tau g-\mu_\tau\|_2^2.
\label{eq:favr_joint}
\end{equation}
Both $Q_\tau$ and $\mu_\tau$ vary with the flow state, allowing channel weights to adapt throughout action generation.

\subsection{Joint Gate Optimization via FISTA}
\label{sec:favr_joint_optimization}

FAVR solves Eq.~\eqref{eq:favr_joint} using FISTA-based joint optimization~\cite{beck2009monotonefista, beck2009fista}. Define $B_\tau=Q_\tau^\top Q_\tau$ and $s_\tau=Q_\tau^\top\mu_\tau$; the off-diagonal terms of $B_\tau$ couple visual channels. The gradient is
\begin{equation}
\begin{aligned}
\nabla F_\tau(g)&=B_\tau g-s_\tau\\
&=x_\tau\odot W^\top
\left[W(x_\tau\odot g)-\mu_\tau\right].
\end{aligned}
\label{eq:favr_gradient}
\end{equation}
It uses products with the frozen head without constructing dense $B_\tau$, while box projection is element-wise clipping. 
With $L_\tau\ge\|Q_\tau\|_2^2$, a feasible initialization $g^{(0)}$, $y^{(0)}=g^{(0)}$, and $\omega_0=1$, FISTA iterates
\begin{equation}
\begin{aligned}
g^{(k+1)}
&=\operatorname{clip}_{[0,1]^d}\!\left(
y^{(k)}-\frac{\nabla F_\tau(y^{(k)})}{L_\tau}
\right),\\
\omega_{k+1}
&=\frac{1+\sqrt{1+4\omega_k^2}}{2},\\
y^{(k+1)}
&=g^{(k+1)}+
\frac{\omega_k-1}{\omega_{k+1}}
\bigl(g^{(k+1)}-g^{(k)}\bigr).
\end{aligned}
\label{eq:favr_fista}
\end{equation}

Here $k$ indexes FISTA iterations. After $N_{\mathrm{iter}}$ iterations at fixed $\tau$, the resulting gate $\widehat g_\tau$ is used to predict the corrected velocity $v_\tau^0+Q_\tau\widehat g_\tau$, supervised by valid-coordinate MSE against $u$. The residual loss updates $\phi$ with detached inputs, while the action loss holds $\phi$ fixed and propagates through the predictor inputs and solver steps to train the visual interface. Both pretrained backbones remain frozen. Each FISTA iteration costs $O(md)$ per action position, giving $O(N_{\mathrm{iter}}md)$ for the full solver. A gate-regularized extension is derived in Appendix~\ref{app:favr_gate_regularization}.

\begin{figure}[t]
    \centering
    \includegraphics[width=\linewidth]{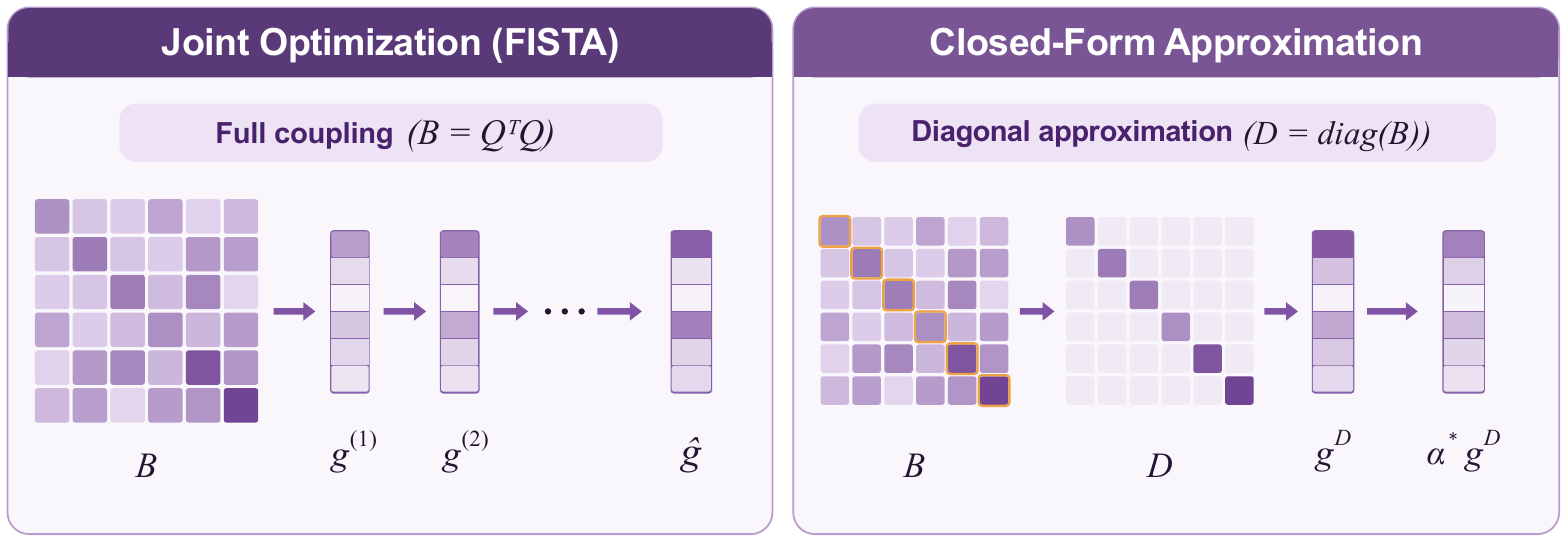}
    \caption{Comparison of joint and closed-form gate optimization. FISTA iteratively optimizes channel weights while retaining cross-channel coupling, whereas the closed-form approximation first computes a diagonal gate $g^D$ and then applies analytic rescaling to obtain $g^{\mathrm{CF}}=\alpha^\star g^D$.}
    \label{fig:favr_gate_comparison}
\end{figure}

\subsection{An Approximate Closed-Form Gate}
\label{sec:favr_closed_form}

Joint optimization in FAVR accounts for channel coupling but requires iterative updates at every flow evaluation. To reduce this computational overhead, we introduce FAVR-CF as a lower-cost approximation that trades a modest reduction in task performance for improved inference efficiency. As illustrated in Fig.~\ref{fig:favr_gate_comparison}, FAVR-CF approximates the joint solution through two non-iterative steps: diagonal approximation and analytic rescaling.

Omitting $\tau$, let $B=Q^\top Q$ and $s=Q^\top\mu$. We retain the diagonal part $D$ of $B$, with $D_{jj}=B_{jj}=\|Q_{:j}\|_2^2$, while omitting off-diagonal interactions. The resulting bounded diagonal solution is
\begin{equation}
g_j^D=
\begin{cases}
\operatorname{clip}_{[0,1]}\!\left(
\dfrac{s_j}{D_{jj}}
\right),&D_{jj}>0,\\[4pt]
0,&D_{jj}=0.
\end{cases}
\label{eq:favr_diagonal}
\end{equation}
Here $s_j$ measures alignment with the residual and $D_{jj}$ represents the squared velocity contribution of channel $j$.

\begin{table*}[t]
\centering
\caption{Success rates (\%) on LIBERO-Plus across seven perturbation categories. Avg.\ is computed over all 10,030 evaluation scenarios.}
\label{tab:favr_liberoplus}
\small                                
\setlength{\tabcolsep}{4pt}
\renewcommand{\arraystretch}{1.15}
\begin{tabular*}{\textwidth}{@{\extracolsep{\fill}}lcccccccc@{}}
\toprule
Method & Camera & Robot & Language & Light &
Background & Noise & Layout & Avg. \\
\midrule
VLA-JEPA~\cite{sun2026vlajepa}
& 63.30 & 67.10 & 85.40 & 95.60 & 93.60 & 66.30 & 85.10 & 78.00 \\
$\pi_{0.5}$ (Frozen)
& 68.36 & 78.58 & 86.34 & 95.79 & 95.91 & 86.45 & 85.25 & 84.22 \\
JEPA-VLA~\cite{jepavla}
& 67.23 & 76.52 & 85.82 & 93.43 & 91.54 & 81.07 & 84.66 & 81.96 \\
PVI~\cite{pvi}
& 69.17 & 77.48 & 87.83 & 95.18 & 92.19 & 81.07 & 85.31 & 83.10 \\
\textbf{FAVR-CF}
& \textbf{69.17} & \textbf{79.10} & \textbf{86.92} & \textbf{96.76} & \textbf{95.35} & \textbf{85.63} & \textbf{85.90} & \textbf{84.55} \\
\textbf{FAVR}
& \textbf{70.42} & \textbf{79.87} & \textbf{88.03} & \textbf{96.41} & \textbf{96.00 }& \textbf{85.82} & \textbf{85.64} & \textbf{85.05} \\
\bottomrule
\end{tabular*}
\end{table*}

To calibrate the combined correction $c^D=Qg^D$, we minimize the regularized one-dimensional objective
$\|\alpha c^D-\mu\|_2^2/2+
\varepsilon_{\mathrm{num}}\alpha^2/2$
over $\alpha\in[0,1]$, yielding
\begin{equation}
\begin{aligned}
g^{\mathrm{CF}}&=\alpha^\star g^D,\\
\alpha^\star
&=\operatorname{clip}_{[0,1]}\!\left(
\frac{\mu^\top c^D}
{\|c^D\|_2^2+\varepsilon_{\mathrm{num}}}
\right).
\end{aligned}
\label{eq:favr_rescale}
\end{equation}
This rescaling adjusts the magnitude of the combined velocity correction while preserving relative channel weights.

Using cached action-head column norms, FAVR-CF computes the gate without explicitly forming $Q$ or $B$, requiring $O(md)$ arithmetic and $O(d+m)$ additional space per action position, compared with $O(N_{\mathrm{iter}}md)$ for FISTA. Its approximation bound is provided in Appendix~\ref{app:favr_cf_proof}, and its performance and latency are evaluated in Sec.~\ref{sec:ablation}.


\section{Experiments}
\label{sec:experiments}

\subsection{Experimental Setup}
\label{sec:experimental_setup}


\begin{table}[t]
\centering
\caption{Success rates (\%) on LIBERO.}
\label{tab:favr_libero}
\small
\setlength{\tabcolsep}{3.5pt}
\renewcommand{\arraystretch}{1.15}
\begin{tabular*}{\columnwidth}{@{\extracolsep{\fill}}lccccc@{}}
\toprule
Method & Spatial & Object & Goal & Long & Avg. \\
\midrule
OpenVLA-OFT~\cite{oft}
& 97.6 & 98.4 & 97.9 & 94.5 & 97.10 \\
MemoryVLA~\cite{memoryvla}
& 98.4 & 98.4 & 96.4 & 93.4 & 96.65 \\
VLA-JEPA~\cite{sun2026vlajepa}
& 96.2 & 99.6 & 97.2 & 95.8 & 97.20 \\
$\pi_{0.5}$ (Frozen)
& 98.0 & 98.2 & 97.0 & 96.4 & 97.40 \\
JEPA-VLA~\cite{jepavla}
& 98.8 & 99.4 & 97.8 & 95.2 & 97.80 \\
PVI~\cite{pvi}
& 98.4 & 100.0 & 97.2 & 94.6 & 97.55 \\
\textbf{FAVR-CF}
& \textbf{99.2} & \textbf{99.4} & \textbf{97.6} & \textbf{96.8} & \textbf{98.25} \\
\textbf{FAVR}
& \textbf{98.8} & \textbf{100.0} & \textbf{98.4} & \textbf{97.2} & \textbf{98.60} \\
\bottomrule
\end{tabular*}
\end{table}

\noindent\textbf{Benchmarks.}
We evaluate FAVR on LIBERO~\cite{libero}, LIBERO-Plus~\cite{liberoplus}, and RoboTwin 2.0~\cite{robotwin}. LIBERO contains four suites (Spatial, Object, Goal, and Long) with 10 language-conditioned tasks each. We report suite-wise and average success rates over 50 episodes per task. LIBERO-Plus evaluates 10,030 scenarios across seven perturbation categories: camera viewpoints, robot initial states, language instructions, lighting, background textures, sensor noise, and object layout. RoboTwin 2.0 comprises 50 bimanual manipulation tasks. We evaluate both Clean-to-Clean (C2C) and Clean-to-Randomized (C2R) settings, reporting task-averaged success rates over 100 rollouts per task in each setting.

\noindent\textbf{Models.}
We instantiate FAVR on LIBERO-finetuned $\pi_{0.5}$~\cite{pi05} for LIBERO and LIBERO-Plus, and a RoboTwin-post-trained 4B LingBot-VLA~\cite{lingbot} for RoboTwin 2.0. We use a frozen V-JEPA 2.1 ViT-B video encoder~\cite{vjepa21} for temporal visual conditioning, while keeping both VLA backbones frozen. Only the temporal visual interface and velocity-residual predictor are trained, introducing 2.190M and 1.736M parameters, respectively.

In simulation, FAVR adapters are trained for 5,000 AdamW steps at a learning rate of $5\times10^{-5}$, with effective batch sizes of 32 for LIBERO and LIBERO-Plus and eight for RoboTwin. We train separate LIBERO and LIBERO-Plus adapters, the latter using 14,347 perturbed demonstrations, while RoboTwin uses 50 clean demonstrations per task. Evaluation uses 10 flow steps to predict 50-action chunks, executing the first 10 actions on LIBERO and LIBERO-Plus and all 50 on RoboTwin. FAVR uses 10-frame histories from one external camera on LIBERO and LIBERO-Plus, and from the head and two wrist cameras on RoboTwin. We denote the FISTA-based joint method as FAVR and its approximate closed-form variant as FAVR-CF. FAVR uses 8 FISTA iterations on LIBERO and 128 on LIBERO-Plus and RoboTwin 2.0. Residual-loss weights are $1.0$ for $\pi_{0.5}$ and $0.5$ for LingBot-VLA. Further implementation details are provided in Appendix~\ref{app:favr_implementation}.

\begin{table}[t]
\centering
\caption{Success rates (\%) on RoboTwin 2.0. C2C and C2R denote Clean-to-Clean and Clean-to-Randomized evaluation, respectively.}
\label{tab:favr_robotwin}
\small
\begin{tabular}{lcc}
\toprule
Method & C2C & C2R\\
\midrule
$\pi_{0.5}$~\cite{lingbot}
& 82.74 & 76.76\\
LingBot-VLA 
& 87.12 & 86.88\\
\textbf{FAVR}
& \textbf{90.04} & \textbf{89.38}\\
\bottomrule
\end{tabular}
\end{table}

\begin{figure*}[t]
    \centering
    \includegraphics[width=0.8\textwidth]{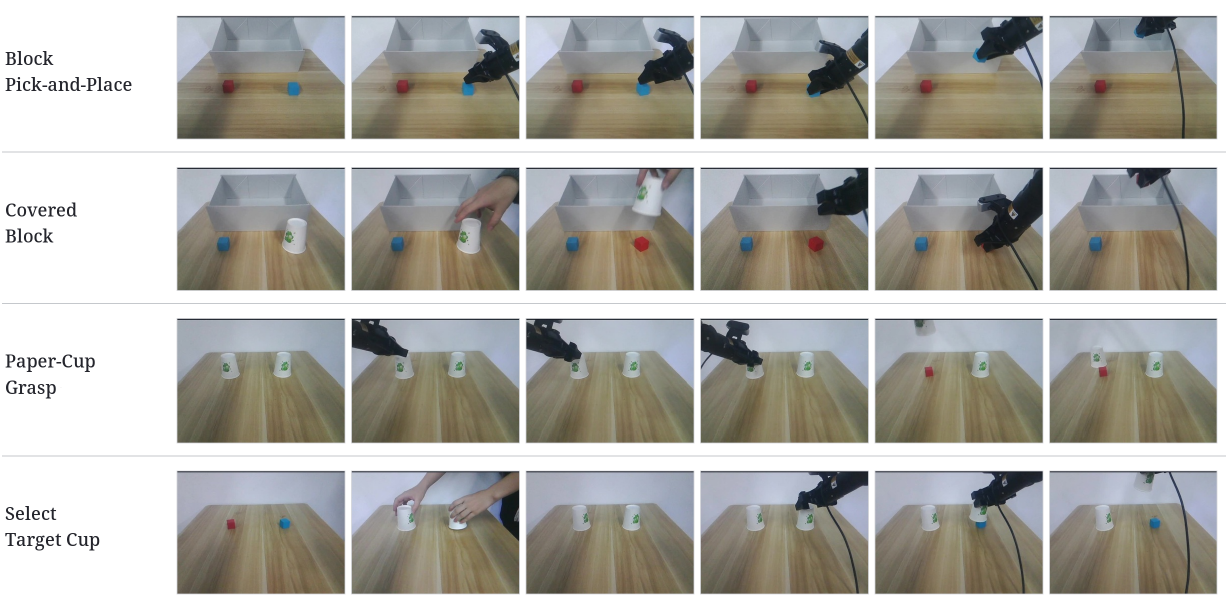}
    \caption{Representative Cobot Magic rollouts: block pick-and-place, paper-cup grasping, covered-block retrieval, and history-conditioned cup selection. The last two tasks require visual history and separate observation from execution.}
    \label{fig:favr_real_rollouts}
\end{figure*}

\noindent\textbf{Baselines.}
We compare against the corresponding frozen VLA backbones and visual augmentation methods JEPA-VLA~\cite{jepavla} and PVI~\cite{pvi}. We further compare with some representative VLA methods~\cite{oft, memoryvla}.

\subsection{Main Results}
\label{sec:main_results}

Tables~\ref{tab:favr_liberoplus}--\ref{tab:favr_robotwin} compare FAVR, JEPA-VLA, PVI with frozen backbones, and representative VLA policies across three benchmarks.

\noindent\textbf{LIBERO.}
Table~\ref{tab:favr_libero} shows that FAVR improves the frozen $\pi_{0.5}$ from 97.40\% to 98.60\% across all four suites, achieving the highest average success rate in Table~\ref{tab:favr_libero}. It trains only 2.190M parameters, versus 24.129M for JEPA-VLA and 444.445M for PVI. FAVR-CF reaches 98.25\%, improving over the frozen baseline by 0.85 pp while trailing joint optimization by 0.35 pp. The largest suite-level gains arise on Object (+1.80 pp) and Goal (+1.40 pp), where the frozen policy already exceeds 97\% success.

\noindent\textbf{LIBERO-Plus.}
Table~\ref{tab:favr_liberoplus} shows that FAVR improves LIBERO-Plus success from 84.22\% to 85.05\%, with gains in six of seven perturbation categories. The largest gains arise from camera viewpoints (+2.06 pp), language instructions (+1.69 pp), and robot initialization (+1.29 pp), indicating benefits under both visual and manipulation changes. FAVR-CF achieves 84.55\%: 0.33 pp above the frozen base and only 0.50 pp below FAVR. Compared with JEPA-VLA (81.96\%) and PVI (83.10\%), FAVR makes better use of auxiliary video conditioning under these perturbations. Sensor noise is the one category without improvement, so the gains are not uniform across all shifts.

\noindent\textbf{RoboTwin 2.0.}
On RoboTwin 2.0 (Table~\ref{tab:favr_robotwin}), FAVR improves frozen LingBot-VLA from 87.12\% to 90.04\% in Clean-to-Clean (C2C) and from 86.88\% to 89.38\% in Clean-to-Randomized (C2R). Using 1.736M trainable parameters and three-view visual history, these gains extend FAVR to bimanual manipulation and randomized conditions without modifying the action expert.


\begin{table}[t]
\centering
\caption{Real-world manipulation results on Cobot Magic. Each task uses 35
demonstrations and 20 evaluation trials.}
\label{tab:favr_real_robot}
\small
\setlength{\tabcolsep}{4pt}
\renewcommand{\arraystretch}{1.15}
\begin{tabular*}{\columnwidth}{@{\extracolsep{\fill}}lccc@{}}
\toprule
Task & Initial FT & Continued FT & FAVR(Ours) \\
\midrule
Block PnP
& 10/20 & 12/20 & \textbf{15/20} \\
Paper-cup grasping
& 8/20 & 11/20 & \textbf{13/20} \\
\bottomrule
\end{tabular*}
\end{table}

\begin{table*}[t]
\centering
\caption{Ablation results on five RoboTwin 2.0 tasks. All adapted variants are trained for 1K steps and evaluated over 50 trials per task. The frozen baseline uses previously obtained evaluation results.}
\label{tab:favr_ablation}
\small
\setlength{\tabcolsep}{6pt}
\begin{tabular}{lcccccc}
\toprule
Method
& Hanging mug
& Can/pot
& Bell
& Bread/skillet
& Can/basket
& Avg. \\
\midrule
Frozen baseline       & 47\% & 72\% & 77\% & 84\% & 64\% & 68.8\% \\
Non-flow-aware gate   & 50\% & 66\% & 88\% & 88\% & 68\% & 72.0\% \\
Learned gate          & 42\% & 74\% & 70\% & 90\% & 72\% & 69.6\% \\
Direct output         & 50\% & 86\% & 90\% & 90\% & 78\% & 78.8\% \\
Hidden-only           & 52\% & 82\% & 86\% & 92\% & 86\% & 79.6\% \\
Window-wise reset     & 52\% & 86\% & 88\% & 90\% & 72\% & 77.6\% \\
Static pooling        & 40\% & 78\% & 76\% & 88\% & 62\% & 68.8\% \\
\midrule
\textbf{FAVR (Ours)}
& \textbf{58\%} & \textbf{86\%} & \textbf{94\%}
& \textbf{94\%} & \textbf{78\%} & \textbf{82.0\%} \\
\bottomrule
\end{tabular}
\end{table*}

\subsection{Real-World Experiments}
\label{sec:real_world}

We further evaluate FAVR on Cobot Magic, a dual-arm platform based on Mobile ALOHA~\cite{cobotmagic}. We consider four tasks, collecting 35 demonstrations per task. Representative execution sequences are shown in Fig.~\ref{fig:favr_real_rollouts}.

\noindent\textbf{Standard manipulation.}
We evaluate color-conditioned block pick-and-place and left/right paper-cup grasping, the latter requiring more precise positioning and grasp control. For each task, we first fine-tune a pretrained bimanual $\pi_{0.5}$ policy, then compare continued fine-tuning with FAVR adaptation from the same checkpoint under matched additional training budgets. During FAVR adaptation, both $\pi_{0.5}$ and V-JEPA remain frozen.

As shown in Table~\ref{tab:favr_real_robot}, the initial block policy achieves 10/20 successes after 5,000 steps. With 3,000 additional steps, continued fine-tuning reaches 12/20, whereas FAVR achieves 15/20. For paper-cup grasping, the initial policy reaches 8/20 after 15,000 steps; with another 5,000 steps, continued fine-tuning achieves 11/20, compared with 13/20 for FAVR.

\noindent\textbf{History-dependent manipulation.}
TTT has been explored as a memory mechanism for sequential modeling~\cite{ttt}, with recent robotic approaches employing recurrent states to capture interaction histories~\cite{robottt, tttrm, t2mem}. Since FAVR also employs a TTT-based temporal visual interface, we further examine its ability to support memory-dependent manipulation through two tasks requiring information unavailable in the current observation.

In covered-block retrieval, the robot must place a previously covered block into a container after its cover is removed. In history-conditioned cup selection, the robot must select the cup concealing a block of the instructed color, whose location was visible before covering. Both tasks separate observation from execution, with the robot remaining stationary during observation to prevent target information from leaking through its own movements. Without action suppression, small arm movements during the observation phase can reveal the selected side through proprioception, allowing the policy to bypass the intended use of visual history.

After 5,000 training steps, FAVR correctly selects the target in 7/10 trials on each task. While exceeding the 50\% chance level for binary selection, the improvement remains modest. These results suggest that the TTT state can provide useful temporal context, while its primary role in FAVR is to support visual conditioning rather than serve as a dedicated memory module.

\noindent\textbf{Dependence on pretrained manipulation capability.}
To test dependence on the base policy, we freeze an unadapted bimanual $\pi_{0.5}$ and V-JEPA checkpoint and train only FAVR for 5,000 steps on block pick-and-place demonstrations. The resulting arms repeatedly move above the workspace without reliably approaching or grasping the blocks. Unlike adaptation from a task-fine-tuned checkpoint, this does not establish stable manipulation: FAVR improves visual conditioning on a base policy that already possesses relevant task-level capabilities.

\subsection{Ablation Studies}
\label{sec:ablation}

We investigate the key design choices of FAVR on five RoboTwin 2.0 tasks: \texttt{hanging\_mug}, \texttt{move\_can\_pot}, \texttt{click\_bell}, \texttt{place\_bread\_skillet}, and \texttt{place\_can\_basket}. All variants are trained for 1,000 steps with a batch configuration of $2\times4$ and a learning rate of $5\times10^{-5}$, and evaluated over 50 trials per task. The frozen LingBot-VLA baseline achieves 68.8\% average success (Table~\ref{tab:favr_ablation}).

\begin{figure*}[t]
    \centering
    \includegraphics[width=0.75\textwidth]{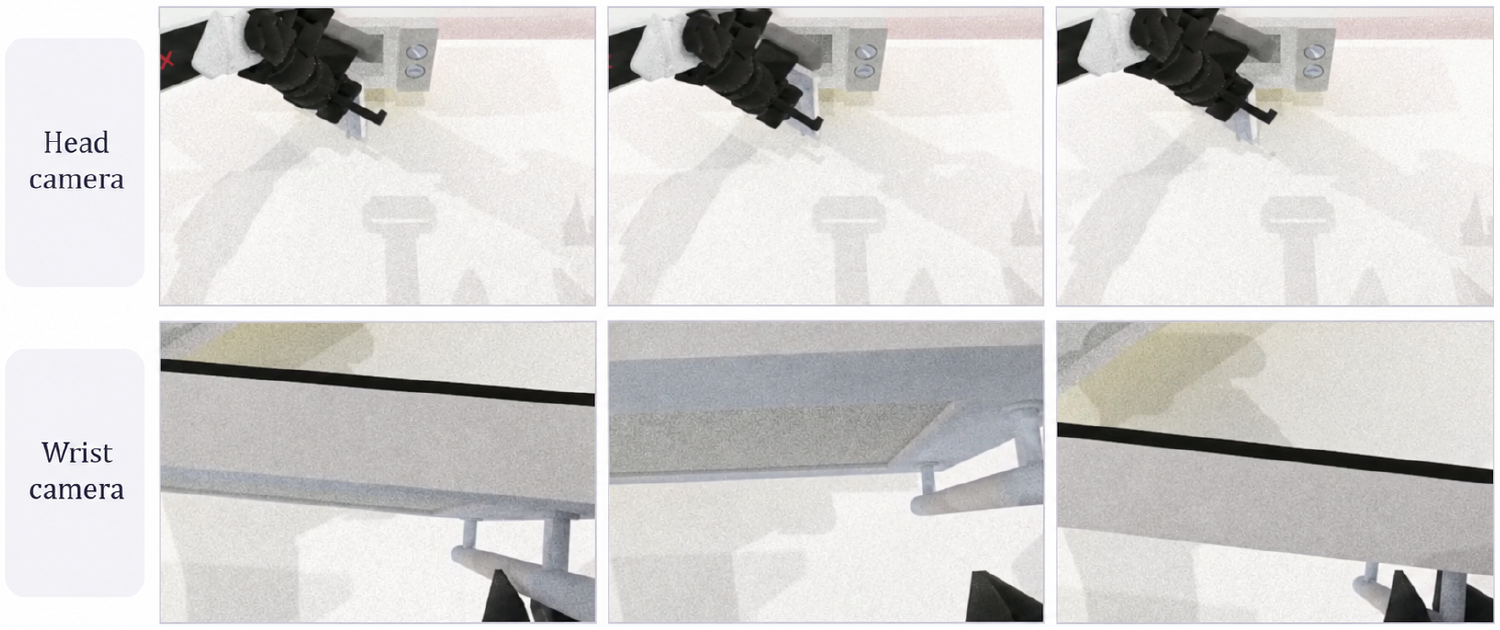}
    \caption{Benefits of multi-view visual conditioning in \texttt{open\_microwave}. Synchronized rollout frames show that gripper slippage is difficult to identify from the head camera but visible from the wrist camera.}
    \label{fig:favr_microwave_views}
\end{figure*}

\noindent\textbf{Flow-aware gating.}
As illustrated in Fig.~\ref{fig:flow_aware_motivation}, FAVR adapts visual channel weights to the current flow state. Since the predicted velocity evolves throughout flow integration, channel weights selected at the initial step may no longer provide suitable corrections at later steps. To examine this, we compare FAVR with a variant that reuses the initial gate across all flow steps while retaining the action-conditioned visual readout. Recomputing the gate increases average success from 72.0\% to 82.0\%. On \texttt{move\_can\_pot}, success rises from 66\% to 86\%, illustrating how a gate selected at one flow state can become less suitable as the velocity field evolves. This improvement highlights the importance of adapting visual corrections to the evolving action-generation state.

\noindent\textbf{Injection position and action-head geometry.}
A key property of FAVR is that the visual correction passes directly through the frozen affine action head, yielding the exact mapping
$\Delta v_\tau=W\operatorname{diag}(x_\tau)g_\tau$.
This enables gate optimization directly in velocity space. At earlier injection positions, however, subsequent nonlinear layers invalidate this mapping. Applying the same head-derived optimization at layers 12, 24, and 36 yields success rates of 67.2\%, 66.8\%, and 65.6\%, respectively, all below the frozen baseline. We therefore also evaluate learned gates that implicitly model the downstream nonlinear response. These improve success to 75.2\%, 79.6\%, and 74.4\%, but remain below FAVR's 82.0\% (see Appendix~\ref{app:favr_injection}). Under the limited adaptation budget, final-head injection is preferable because the known channel-to-velocity mapping permits direct gate optimization instead of learning the downstream nonlinear response.

\noindent\textbf{Correction mechanisms.}
We test whether the gains arise from velocity-residual alignment rather than simply adding a correction branch. We compare three alternatives: \emph{Learned Gate}, which directly predicts channel weights; \emph{Direct Output}, which predicts bounded velocity corrections without gated feature injection; and \emph{Hidden-Only}, which removes the temporal video-state interface and relies solely on VLA hidden features. Their average success rates are 69.6\%, 78.8\%, and 79.6\%, respectively, compared with 82.0\% for FAVR.

\noindent\textbf{Temporal visual state.}
We test whether recurrent visual states outperform independent video windows. Resetting TTT after each window reduces average success from 82.0\% to 77.6\%, while replacing recurrent updates with static feature pooling further reduces success to 68.8\%. These results indicate that accumulating observations across windows provides more useful visual context than independent window processing or static aggregation.

\noindent\textbf{Video representation and multi-view conditioning.}
We first examine the contribution of video-pretrained representations. On LIBERO, replacing V-JEPA with I-JEPA reduces average success from 98.6\% to 97.8\%, with the largest difference on LIBERO-Long (97.2\% versus 94.8\%). This suggests that video-pretrained features are particularly useful for manipulation tasks involving extended temporal interactions.

We further investigate whether additional camera views improve the visual information available to V-JEPA. On the full 50-task RoboTwin 2.0 benchmark, conditioning V-JEPA on the head camera alone yields 88.48\% average success, whereas incorporating the head and two wrist cameras increases success to 90.04\%. The VLA policy retains its three-view observations in both settings.

The \texttt{open\_microwave} task illustrates why multi-view conditioning can be valuable (Fig.~\ref{fig:favr_microwave_views}). During manipulation, the gripper may slip from the handle, requiring the robot to recover its grasp before continuing. Such contact changes are difficult to distinguish from the head camera but clearly visible from the wrist camera. With head-only video conditioning, the auxiliary visual representation may miss this local interaction change and continue to favor nominal task progression. Incorporating wrist observations provides additional evidence of the actual contact state, allowing visual corrections to better reflect the need for recovery. On this task, three-view conditioning improves success from 45\% to 64\% (+19 pp). This example helps explain why multi-view video representations can substantially benefit manipulation tasks requiring precise contact monitoring and recovery.

\noindent\textbf{Inference cost.}
V-JEPA processes ten-frame windows during execution, with the final update overlapping VLA inference at replanning. Averaged over 100 RoboTwin 2.0 calls (50-action chunks, 10 flow steps), inference latency is 1117.7\,ms for the frozen baseline, 1134.3\,ms for FAVR-CF, and 1487.3\,ms for FAVR. FAVR-CF thus reduces latency by 23.7\% relative to FAVR with only 1.5\% overhead over the frozen baseline, while trailing FAVR by 0.35 and 0.50 percentage points on LIBERO and LIBERO-Plus, respectively.

\section{Conclusion}
\label{sec:conclusion}

We presented FAVR, a lightweight framework for integrating pretrained video representations into frozen vision-language-action policies. FAVR constructs a compact temporal visual state through TTT and performs flow-aware visual residual alignment by exploiting the affine structure of the frozen action head. This formulation enables direct gate optimization in velocity space, for which we develop a FISTA-based joint solver and an approximate closed-form alternative, FAVR-CF. Experiments on LIBERO, LIBERO-Plus, RoboTwin 2.0, and real-world manipulation demonstrate consistent improvements with few trainable parameters, while FAVR-CF offers a favorable trade-off between performance and inference efficiency. Future work may extend velocity-residual alignment to more general action mappings and explore joint adaptation of visual representations and action policies.

\FloatBarrier
\clearpage
\appendix
\setcounter{figure}{0}\renewcommand{\thefigure}{S\arabic{figure}}
\setcounter{table}{0}\renewcommand{\thetable}{S\arabic{table}}
\setcounter{equation}{0}\renewcommand{\theequation}{S\arabic{equation}}
\section{Implementation Details}
\label{app:favr_implementation}

We distinguish environment time $t$, which advances with observations, from flow time $\tau$, which indexes velocity evaluations within one action chunk. The pretrained VLA and V-JEPA~2.1 encoders remain frozen throughout FAVR adaptation.

\subsection{Video encoding and temporal state}
\label{app:video_state}
FAVR encodes consecutive ten-frame RGB windows at $384\times384$ resolution using frozen V-JEPA~2.1 ViT-B~\cite{vjepa21} (\texttt{vjepa2\_1\_vit\_base\_384}, with \texttt{ema\_encoder} weights from \texttt{vjepa2\_1\_vitb\_dist\_vitG\_384.pt}). Windows advance by ten observed frames. LIBERO and LIBERO-Plus use one external camera, while RoboTwin~2.0 concatenates video tokens from the head and two wrist cameras into a shared TTT state.

Token-level layer normalization and learned projections produce normalized keys and $\tanh$-bounded values of width $p=128$. The FP32 state $S\in\mathbb R^{p\times p}$ is initialized from learned $S_0$ and updated sequentially. For a block $(K,V)$ containing $n$ tokens, the update is
\begin{equation}
S\leftarrow S-\frac{\eta}{n}K^\top(KS-V),\qquad \eta=\operatorname{sigmoid}(\eta_{\rm raw}).
\label{eq:sup_state_update}
\end{equation}
Tokens within each block use the same incoming state, and updates proceed sequentially across blocks. The state is reset to learned $S_0$ at each episode boundary, retained across observed windows within an episode, and held fixed during the flow integration of one action chunk. Only observed frames contribute to writes; no predicted future image or unobserved frame is used for current action conditioning.

\subsection{Action-conditioned retrieval and trainable modules}
At action token $k$ and flow time $\tau$, the frozen policy's final normalized hidden feature $h_{t,\tau,k}$ forms a query to the committed state $S_t$. Using column vectors, the learned readout is
\begin{equation}
q=\operatorname{norm}(P_Q^\top\operatorname{LN}(h)),\quad
\widetilde x=P_O^\top S_t^\top q+b_O.
\end{equation}
The feature is capped relative to the unmodified policy feature,
\begin{equation}
x=\widetilde x\,\operatorname{clip}_{[0,1]}\!\left(
\frac{\rho\operatorname{sg}(\|h\|_2)}{\|\widetilde x\|_2+\varepsilon_{\rm num}}
\right).
\label{eq:sup_readout_cap}
\end{equation}
We use $\rho=0.1$; $\operatorname{sg}$ denotes stop-gradient and $\varepsilon_{\rm num}>0$ provides numerical stabilization. The trainable interface comprises the initial state, key/value/query/output projections and associated normalization, together with the velocity-residual predictor. The resulting trainable totals are 2.190M for $\pi_{0.5}$ and 1.736M for LingBot-VLA, including the auxiliary modules but excluding both frozen backbones.

The residual predictor $R_\phi$ receives the current action feature $h$, retrieved readout $x$, projected contribution $Wx$, and a flow-time embedding $e(\tau)$. It estimates $\mu\approx u-v^0$ with valid-coordinate MSE, where $u$ is the rectified-flow training target and $v^0$ is the frozen policy velocity. During the residual-predictor update, the predictor inputs and target are detached. During the action-loss update, $\phi$ is held fixed while gradients propagate through its inputs, the finite gate-solver steps, and the visual interface. The \emph{base} VLA and V-JEPA weights remain frozen in both paths. The residual predictor has three linear layers with SiLU activations after the first two. For LingBot-VLA, its action hidden size, flow-time embedding size, and output velocity dimension are $768$, $768$, and $14$, giving an MLP of $2318\!\rightarrow\!512\!\rightarrow\!256\!\rightarrow\!14$. For $\pi_{0.5}$, these dimensions are $1024$, $1024$, and $7$, giving $3079\!\rightarrow\!512\!\rightarrow\!256\!\rightarrow\!7$. The concatenated inputs are $[h,x,Wx,e(\tau)]$. The residual auxiliary loss coefficient is $0.5$ for LingBot-VLA and $1.0$ for $\pi_{0.5}$.

\subsection{Causal state scheduling at inference}
For each environment, the inference sequence is: (i) encode newly observed frames when a complete ten-frame window is available; (ii) commit the TTT updates; (iii) generate a new action chunk using the fixed committed state and a flow-dependent action query; and (iv) incorporate subsequently observed frames after execution. The $\pi_{0.5}$ protocols predict 50 actions and execute the first ten before replanning, whereas RoboTwin executes all 50; thus a 50-action RoboTwin interval can incorporate five consecutive observed ten-frame windows before the next plan. Replanning and visual-state updates operate on different clocks.

\subsection{Benchmark-specific training configurations}
Table~\ref{tab:sup_training} collects the settings associated with the main benchmarks. The two LIBERO-family adapters are trained separately; their initial backbone is the LIBERO-finetuned $\pi_{0.5}$ checkpoint. RoboTwin uses the 4B LingBot-VLA checkpoint post-trained on RoboTwin. The standard-LIBERO recipe samples four trajectories with eight consecutive supervision points; the RoboTwin recipe uses two trajectories with four. 

\begin{table*}[t]
\centering\small
\caption{\textbf{Main benchmark training and evaluation configurations.} Steps are optimizer updates and the evaluation horizon is written as predicted/executed actions. Views refer to the auxiliary V-JEPA branch; the original VLA observation interface is retained.}
\label{tab:sup_training}
\setlength{\tabcolsep}{5pt}
\resizebox{\textwidth}{!}{%
\begin{tabular}{lccccccc}
\toprule
Benchmark & Base policy & Steps & LR & Eff. batch & Adapter data & V-JEPA views & FISTA / horizon \\
\midrule
LIBERO & $\pi_{0.5}$ & 5,000 & $5\!\times\!10^{-5}$ & 32 & LIBERO demonstrations & 1 & 8 / 50:10 \\
LIBERO-Plus & $\pi_{0.5}$ & 5,000 & $5\!\times\!10^{-5}$ & 32 & 14,347 perturbed demos & 1 & 128 / 50:10 \\
RoboTwin 2.0 & LingBot-VLA 4B & 5,000 & $5\!\times\!10^{-5}$ & 8 & 50 clean demos/task & 3 & 128 / 50:50 \\
\bottomrule
\end{tabular}%
}
\end{table*}

AdamW training uses a 5\% warmup, cosine learning-rate decay, zero weight decay, and gradient clipping at one. The standard-LIBERO configuration specifies a gate-regularization coefficient of $10^{-3}$. The RoboTwin single-view and three-view configurations both use a residual auxiliary loss coefficient of $0.5$; the $\pi_{0.5}$/LIBERO family uses $1.0$. 

The main five-task ablations use a separate 1,000-update budget, learning rate $5\times10^{-5}$, batch configuration $2\times4$, and 50 trials per task.

\section{Optimization Details and Proofs}
\label{app:favr_optimization}

\subsection{Velocity-space gate objective}
For a fixed action token and flow step, let $h$ denote the frozen policy feature, $x$ the retrieved visual readout, and $g\in[0,1]^d$ the channel gate. The frozen affine action head gives
\begin{equation}
W(h+g\odot x)+b=(Wh+b)+Qg,\qquad Q=W\operatorname{diag}(x).
\label{eq:sup_affine_identity}
\end{equation}
The induced correction is linear in the gate, so matching the predicted velocity residual $\mu$ leads to
\begin{equation}
\min_{g\in[0,1]^d}F(g),\qquad F(g)=\tfrac12\|Qg-\mu\|_2^2.
\label{eq:sup_joint_objective}
\end{equation}
For intermediate-layer injection, subsequent nonlinear transformations alter the effect of the injected feature. The final affine interface instead permits direct box-constrained optimization in velocity space.

\subsection{FISTA implementation}
\label{app:favr_fista_details}
Define $B=Q^\top Q$ and $s=Q^\top\mu$. We evaluate the gradient without constructing $B$:
\begin{equation}
\nabla F(g)=Bg-s=x\odot W^\top\!\bigl(W(x\odot g)-\mu\bigr).
\end{equation}
For $L\ge\|Q\|_2^2$, the accelerated projected iterations are
\begin{align}
g^{(k+1)}&=\operatorname{clip}_{[0,1]^d}\!\left(y^{(k)}-\frac{\nabla F(y^{(k)})}{L}\right),\\
\omega_{k+1}&=\frac{1+\sqrt{1+4\omega_k^2}}2,\nonumber\\
y^{(k+1)}&=g^{(k+1)}+\frac{\omega_k-1}{\omega_{k+1}}(g^{(k+1)}-g^{(k)}).
\end{align}
The implementation uses $L=\max(1.00001\lambda_{\max}(QQ^\top),10^{-12})$. If an accelerated candidate increases $F$, the solver falls back to a projected step from the preceding feasible iterate. Gradients through the visual interface propagate through the executed solver steps. The iteration budgets are eight on LIBERO and 128 on LIBERO-Plus and RoboTwin 2.0.

\subsection{Extension to quadratic gate regularization}
\label{app:favr_gate_regularization}
Consider a quadratic penalty with coefficient $\lambda\ge0$:
\begin{equation}
F_\lambda(g)=\tfrac12\|Qg-\mu\|_2^2+\tfrac\lambda2\|g\|_2^2,
\qquad g\in[0,1]^d.
\label{eq:sup_gate_regularized}
\end{equation}
Its gradient and a valid FISTA step-size bound are
\begin{equation}
\nabla F_\lambda(g)=Q^\top(Qg-\mu)+\lambda g,\qquad
L_\lambda\ge\|Q\|_2^2+\lambda.
\end{equation}
In the diagonal surrogate $D=\operatorname{Diag}(Q^\top Q)$, the solution becomes
\begin{equation}
g_{\lambda,j}^{D}=\operatorname{clip}_{[0,1]}\!\left(\frac{s_j}{D_{jj}+\lambda}\right),
\qquad s=Q^\top\mu,
\end{equation}
with $g_{\lambda,j}^{D}=0$ if the denominator vanishes. Set $c_\lambda^D=Qg_\lambda^D$. Including the quadratic gate penalty in scalar rescaling yields
\begin{align}
\min_{0\le\alpha\le1}\;&\tfrac12\|\alpha c_\lambda^D-\mu\|_2^2
+\tfrac\lambda2\alpha^2\|g_\lambda^D\|_2^2
+\tfrac{\varepsilon_{\rm num}}2\alpha^2,\nonumber\\
\alpha_\lambda^*&=\operatorname{clip}_{[0,1]}\!\left(
\frac{\mu^\top c_\lambda^D}{\|c_\lambda^D\|_2^2+
\lambda\|g_\lambda^D\|_2^2+\varepsilon_{\rm num}}\right).
\label{eq:sup_reg_scalar}
\end{align}
The diagonal-solution argument below also gives, for any minimizer $g_\lambda^*$ of $F_\lambda$,
\begin{equation}
0\le F_\lambda(\alpha_\lambda^*g_\lambda^D)-F_\lambda(g_\lambda^*)
\le\|Eg_\lambda^D\|_1+\tfrac{\varepsilon_{\rm num}}2,
\quad E=Q^\top Q-D.
\label{eq:sup_reg_bound}
\end{equation}
Setting $\lambda=0$ recovers the FAVR-CF formulation used in the main manuscript.

\subsection{FAVR-CF derivation and approximation bound}
\label{app:favr_cf_proof}
Decompose $B=Q^\top Q=D+E$, where $D_{jj}=\|Q_{:j}\|_2^2$ and $E$ contains the off-diagonal terms; let $s=Q^\top\mu$. Omitting $E$ gives the separable objective
\begin{equation}
F_D(g)=\tfrac12g^\top Dg-s^\top g+\tfrac12\|\mu\|_2^2,
\end{equation}
whose exact box-constrained solution is
\begin{equation}
g_j^D=\begin{cases}
\operatorname{clip}_{[0,1]}(s_j/D_{jj}),&D_{jj}>0,\\
0,&D_{jj}=0.
\end{cases}
\label{eq:sup_diag_rule}
\end{equation}
If $D_{jj}=0$, column $Q_{:j}$ vanishes and $s_j=0$, so zero is a valid minimizer. With $c^D=Qg^D$, FAVR-CF solves the stabilized scalar problem
\begin{equation}
\min_{0\le\alpha\le1}\;\tfrac12\|\alpha c^D-\mu\|_2^2+
\tfrac{\varepsilon_{\rm num}}2\alpha^2.
\label{eq:sup_cf_scalar}
\end{equation}
Differentiating and projecting the stationary point onto $[0,1]$ yields
\begin{equation}
\alpha^*=\operatorname{clip}_{[0,1]}\!\left(
\frac{\mu^\top c^D}{\|c^D\|_2^2+\varepsilon_{\rm num}}\right),\qquad
g^{\rm CF}=\alpha^*g^D.
\label{eq:sup_cf_rescale}
\end{equation}

\paragraph{Approximation bound.}
Let $g^*$ minimize $F$ over $[0,1]^d$. By the first-order optimality of $g^D$ for $F_D$,
$(Dg^D-s)^\top(g^*-g^D)\ge0$. Convexity of $F$ therefore gives
\begin{align}
F(g^D)-F(g^*)&\le\nabla F(g^D)^\top(g^D-g^*)\nonumber\\
&\le(Eg^D)^\top(g^D-g^*)\le\|Eg^D\|_1.
\label{eq:sup_diagonal_error}
\end{align}
Since $\alpha^*$ minimizes the regularized scalar objective and $\alpha=1$ is feasible,
$F(g^{\rm CF})+\varepsilon_{\rm num}(\alpha^*)^2/2\le F(g^D)+\varepsilon_{\rm num}/2$.
Combining these inequalities and using $\|g^D\|_2\le\sqrt d$ gives
\begin{equation}
\boxed{0\le F(g^{\rm CF})-F(g^*)\le
\|Eg^D\|_1+\tfrac{\varepsilon_{\rm num}}2
\le d\|E\|_2+\tfrac{\varepsilon_{\rm num}}2.}
\label{eq:sup_cf_bound}
\end{equation}

\paragraph{Matrix-free computation.}
For a frozen action head, precompute the action-head column norms and evaluate
\begin{equation}
s=x\odot(W^\top\mu),\qquad D_{jj}=x_j^2\|W_{:j}\|_2^2,\qquad
c^D=W(x\odot g^D).
\end{equation}
The closed-form gate requires $O(md)$ arithmetic and $O(m+d)$ extra workspace per action token; $N_{\mathrm{iter}}$ FISTA steps require $O(N_{\mathrm{iter}}md)$ arithmetic after step-size setup.

\section{Benchmark Protocols and Detailed Results}
\label{app:favr_benchmarks}

\subsection{Evaluation units and reported metrics}
\textbf{LIBERO.} We evaluate the Spatial, Object, Goal, and Long suites~\cite{libero}, each containing ten tasks. Each task has 50 evaluation episodes (500 per suite, 2,000 overall). Suite and overall success rates are equal-weight task averages. The action policy predicts 50 actions and executes the first ten before replanning.

\textbf{LIBERO-Plus.} We evaluate the complete 10,030 perturbed scenarios in seven families: camera, robot initialization, language, lighting, background, sensor noise, and object layout~\cite{liberoplus}. The overall result is pooled across scenarios, rather than the unweighted mean of the seven category percentages. The Plus adapter is trained separately from the standard-LIBERO adapter using 14,347 perturbed demonstrations. Its evaluation also uses the 50/10 action horizon.

\textbf{RoboTwin 2.0.} We evaluate all 50 bimanual tasks~\cite{robotwin}, with 100 rollouts per task in each of the Clean-to-Clean (C2C) and Clean-to-Randomized (C2R) conditions. Task success is averaged equally across the 50 tasks. Adaptation uses 50 \emph{clean} demonstrations per task; randomized conditions are evaluated separately. The policy executes the full predicted 50-action chunk before replanning. The RoboTwin rendering configuration uses ray tracing with 32 samples per pixel, path depth eight, and no denoiser; the policy retains its native action-normalization convention.

\subsection{Clean-to-Randomized scene protocol}
\label{app:favr_c2r_randomization}
RoboTwin C2R applies randomized scene appearance and geometry while evaluating policies adapted using clean demonstrations. Wall and tabletop textures are randomized; each surface can retain its default background with approximately $2\%$ probability (\texttt{clean\_background\_rate=0.02}). Clutter objects are sampled on the tabletop, targeting ten distractors, although placement failures can yield fewer; clutter can also be skipped with approximately $2\%$ probability. The tabletop height is randomly shifted within the configured range of $-3$ to $0\,$cm relative to the default. Directional and point lights are randomized in RGB color and intensity. Dynamic lighting, whose parameters can vary with successive renders, is enabled with $2\%$ probability. Extra head-camera displacement is disabled (parameter $0$).

\subsection{LIBERO task-suite results}
\label{app:favr_libero_counts}
Table~\ref{tab:sup_libero_counts} reports the exact integer successes corresponding to the four-suite rates in the main manuscript.

\begin{table}[t]
\centering\small
\caption{\textbf{LIBERO success counts.} Each suite has 500 trials (2,000 in total).}
\label{tab:sup_libero_counts}
\setlength{\tabcolsep}{3pt}
\begin{tabular}{lrrrrr}
\toprule
Policy & Spat. & Obj. & Goal & Long & Total \\
\midrule
Frozen $\pi_{0.5}$ & 490 & 491 & 485 & 482 & 1,948 \\
FAVR-CF & 496 & 497 & 488 & 484 & 1,965 \\
\textbf{FAVR} & 494 & 500 & 492 & 486 & \textbf{1,972} \\
\bottomrule
\end{tabular}
\end{table}

The overall differences are $+1.20$ percentage points for FAVR and $+0.85$ for FAVR-CF relative to the frozen policy; FAVR-CF trails FAVR by $0.35$ points.

\subsection{Perturbation-level outcomes on LIBERO-Plus}
Table~\ref{tab:sup_plus_breakdown} gives the two local policies' category percentages and their signed differences. The largest positive differences are for viewpoint, instruction, and robot initialization; the sensor-noise category decreases by $0.63$ percentage points.

\begin{table}[t]
\centering\small
\caption{\textbf{LIBERO-Plus perturbation breakdown.} $\Delta$ denotes FAVR minus the corresponding frozen $\pi_{0.5}$ baseline, in percentage points; Avg. pools all 10,030 evaluation scenarios.}
\label{tab:sup_plus_breakdown}
\setlength{\tabcolsep}{3pt}
\begin{tabular}{lrrr}\toprule
Perturbation & Base & FAVR & $\Delta$ (pp)\\\midrule
Camera viewpoint & 68.36 & 70.42 & +2.06 \\
Robot initialization & 78.58 & 79.87 & +1.29 \\
Language instruction & 86.34 & 88.03 & +1.69 \\
Lighting & 95.79 & 96.41 & +0.62 \\
Background texture & 95.91 & 96.00 & +0.09 \\
Sensor noise & 86.45 & 85.82 & -0.63 \\
Object layout & 85.25 & 85.64 & +0.39 \\
\midrule
Avg. & 84.22 & 85.05 & +0.83 \\
\bottomrule\end{tabular}
\end{table}

\subsection{Complete RoboTwin 2.0 task-level results}
\label{app:favr_robotwin_provenance}
Table~\ref{tab:sup_robotwin_final} lists the success rates on all 50 tasks under C2C and C2R. The $\pi_{0.5}$ columns are taken from Table~S9 of LingBot-VLA~\cite{lingbot}; the frozen LingBot-VLA and FAVR columns are evaluated using our RoboTwin setup. Each task has 100 rollouts per condition. FAVR improves the frozen LingBot-VLA average from 87.12\% to 90.04\% under C2C and from 86.88\% to 89.38\% under C2R.

\begin{table*}[t]
\centering\scriptsize
\caption{\textbf{RoboTwin 2.0 success rates (\%) for all 50 tasks.} The $\pi_{0.5}$ columns are reported in LingBot-VLA~\cite{lingbot} (Table S9); Frozen LingBot-VLA and FAVR are our evaluations (100 trials per task per setting). C2C/C2R denote clean/randomized evaluation.}
\label{tab:sup_robotwin_final}
\setlength{\tabcolsep}{3.8pt}
\renewcommand{\arraystretch}{0.99}
\begin{tabular}{lrrrrrr}
\toprule
Task & $\pi_{0.5}$ C2C & Frozen C2C & FAVR C2C & $\pi_{0.5}$ C2R & Frozen C2R & FAVR C2R \\
\midrule
\texttt{adjust\_bottle} & 100 & 100 & 100 & 99 & 98 & 99 \\
\texttt{beat\_block\_hammer} & 96 & 90 & 92 & 93 & 81 & 83 \\
\texttt{blocks\_ranking\_rgb} & 92 & 92 & 90 & 85 & 90 & 93 \\
\texttt{blocks\_ranking\_size} & 49 & 67 & 72 & 26 & 74 & 72 \\
\texttt{click\_alarmclock} & 98 & 78 & 89 & 89 & 86 & 89 \\
\texttt{click\_bell} & 99 & 77 & 92 & 66 & 91 & 94 \\
\texttt{dump\_bin\_bigbin} & 92 & 76 & 83 & 97 & 89 & 91 \\
\texttt{grab\_roller} & 100 & 99 & 100 & 100 & 100 & 100 \\
\texttt{handover\_block} & 66 & 90 & 97 & 57 & 86 & 90 \\
\texttt{handover\_mic} & 98 & 94 & 98 & 97 & 100 & 100 \\
\texttt{hanging\_mug} & 18 & 47 & 64 & 17 & 41 & 49 \\
\texttt{lift\_pot} & 96 & 100 & 100 & 85 & 100 & 100 \\
\texttt{move\_can\_pot} & 51 & 72 & 88 & 55 & 65 & 84 \\
\texttt{move\_pillbottle\_pad} & 84 & 93 & 96 & 61 & 94 & 96 \\
\texttt{move\_playingcard\_away} & 96 & 100 & 100 & 84 & 98 & 100 \\
\texttt{move\_stapler\_pad} & 56 & 71 & 74 & 42 & 64 & 77 \\
\texttt{open\_laptop} & 90 & 95 & 96 & 96 & 100 & 100 \\
\texttt{open\_microwave} & 34 & 82 & 64 & 77 & 60 & 50 \\
\texttt{pick\_diverse\_bottles} & 81 & 76 & 89 & 71 & 72 & 81 \\
\texttt{pick\_dual\_bottles} & 93 & 86 & 98 & 63 & 93 & 95 \\
\texttt{place\_a2b\_left} & 87 & 80 & 86 & 82 & 86 & 87 \\
\texttt{place\_a2b\_right} & 87 & 85 & 85 & 84 & 79 & 79 \\
\texttt{place\_bread\_basket} & 77 & 92 & 90 & 64 & 92 & 95 \\
\texttt{place\_bread\_skillet} & 85 & 84 & 98 & 66 & 83 & 91 \\
\texttt{place\_burger\_fries} & 94 & 99 & 100 & 87 & 97 & 98 \\
\texttt{place\_can\_basket} & 62 & 64 & 78 & 62 & 62 & 68 \\
\texttt{place\_cans\_plasticbox} & 94 & 99 & 100 & 84 & 100 & 99 \\
\texttt{place\_container\_plate} & 99 & 99 & 99 & 95 & 95 & 97 \\
\texttt{place\_dual\_shoes} & 75 & 83 & 82 & 75 & 85 & 92 \\
\texttt{place\_empty\_cup} & 100 & 100 & 100 & 99 & 100 & 100 \\
\texttt{place\_fan} & 87 & 81 & 81 & 85 & 86 & 90 \\
\texttt{place\_mouse\_pad} & 60 & 94 & 93 & 39 & 90 & 93 \\
\texttt{place\_object\_basket} & 80 & 94 & 95 & 76 & 92 & 92 \\
\texttt{place\_object\_scale} & 86 & 87 & 90 & 80 & 88 & 96 \\
\texttt{place\_object\_stand} & 91 & 95 & 97 & 85 & 93 & 95 \\
\texttt{place\_phone\_stand} & 81 & 91 & 93 & 81 & 88 & 96 \\
\texttt{place\_shoe} & 92 & 91 & 91 & 93 & 90 & 93 \\
\texttt{press\_stapler} & 87 & 89 & 89 & 83 & 88 & 90 \\
\texttt{put\_bottles\_dustbin} & 84 & 83 & 90 & 79 & 78 & 78 \\
\texttt{put\_object\_cabinet} & 80 & 86 & 86 & 79 & 85 & 88 \\
\texttt{rotate\_qrcode} & 89 & 90 & 93 & 87 & 89 & 85 \\
\texttt{scan\_object} & 72 & 87 & 89 & 65 & 89 & 87 \\
\texttt{shake\_bottle} & 99 & 100 & 100 & 97 & 98 & 99 \\
\texttt{shake\_bottle\_horizontally} & 99 & 100 & 100 & 99 & 99 & 99 \\
\texttt{stack\_blocks\_three} & 91 & 91 & 89 & 76 & 94 & 93 \\
\texttt{stack\_blocks\_two} & 97 & 98 & 98 & 100 & 99 & 100 \\
\texttt{stack\_bowls\_three} & 77 & 84 & 86 & 71 & 86 & 79 \\
\texttt{stack\_bowls\_two} & 95 & 96 & 93 & 96 & 97 & 97 \\
\texttt{stamp\_seal} & 79 & 80 & 86 & 55 & 92 & 96 \\
\texttt{turn\_switch} & 62 & 69 & 63 & 54 & 62 & 74 \\
\midrule
\textbf{Average} & \textbf{82.74} & \textbf{87.12} & \textbf{90.04} & \textbf{76.76} & \textbf{86.88} & \textbf{89.38} \\
\bottomrule
\end{tabular}
\end{table*}

\section{Additional Results and Ablations}
\label{app:additional_ablations}

\subsection{Common five-task ablation protocol}
The main-text RoboTwin ablations use the task set \texttt{hanging\_mug}, \texttt{move\_can\_pot}, \texttt{click\_bell}, \texttt{place\_bread\_skillet}, and \texttt{place\_can\_basket}. Adapted variants are trained for 1,000 updates with effective batch configuration $2\times4$ and evaluated on 50 trials per task. The frozen LingBot-VLA policy serves as the reference (68.8\% mean success).

\subsection{Flow-dependent gate refitting}
\label{app:favr_flow_gate_ablation}
The Shared Gate and flow-aware FAVR variants were \emph{independently retrained} from the same initialization, producing distinct final checkpoints. Both use the same five-task dataset (50 clean demonstrations per task, 250 total), a common training seed of 7, and 1,000 optimizer updates, each sampling two trajectories with four supervised checkpoints. Their frozen three-view V-JEPA encoder, 10-frame history windows, and TTT state-update procedure are identical. The evaluation pairs the initial scenes and policy noise across the variants: \texttt{demo\_clean}, 50 rollouts per task (250 overall), predicted/executed horizon $50/50$, and ten flow evaluations.

At \emph{every} flow evaluation, both variants query the current TTT state using the current action hidden feature, so the video readout is flow-conditioned in both. FAVR refits the gate at each step; Shared Gate computes the gate at the first step of a chunk and reuses those weights over the subsequent nine steps, recomputing at the next action chunk. Shared Gate uses one additional reference forward pass during training. Mean success increases from 72.0\% with Shared Gate to 82.0\% with flow-aware refitting; \texttt{move\_can\_pot} increases from 66\% to 86\%.

\subsection{Injection depth and alternative correction heads}
\label{app:favr_injection}
At the final affine head the map from gate to correction is exactly $Qg$ at each fixed flow state (Eq.~\eqref{eq:sup_affine_identity}). At earlier action-expert layers, the same final-head-derived quadratic is no longer the exact response after the nonlinear suffix. We compare applying this head-derived optimization at intermediate layers with learning the gate at each location. Table~\ref{tab:sup_early_injection} reports the full five-task results. Learned gates outperform head-derived optimization at every intermediate layer, whereas FAVR at the final affine head obtains the highest mean success.

\begin{table*}[t]
\centering\small
\caption{\textbf{Five-task injection-position experiment.} Success rates (\%) for three earlier injection layers and the final affine interface. All variants are reported with per-task results.}
\label{tab:sup_early_injection}
\setlength{\tabcolsep}{5pt}
\begin{tabular}{llrrrrrr}
\toprule
Position & Gate & Mug & Can/pot & Bell & Bread/skillet & Can/basket & Avg.\\
\midrule
Layer 12 & Head-derived & 42 & 68 & 68 & 94 & 64 & 67.2\\
Layer 12 & Learned & 56 & 74 & 90 & 92 & 64 & 75.2\\
Layer 24 & Head-derived & 46 & 70 & 64 & 86 & 68 & 66.8\\
Layer 24 & Learned & 54 & 80 & 96 & 88 & 80 & 79.6\\
Layer 36 & Head-derived & 54 & 68 & 62 & 88 & 56 & 65.6\\
Layer 36 & Learned & 52 & 80 & 92 & 86 & 62 & 74.4\\
\midrule
Before affine head & \textbf{FAVR} & 58 & 86 & 94 & 94 & 78 & \textbf{82.0}\\
\bottomrule
\end{tabular}
\end{table*}

We additionally contrast three correction mechanisms. \emph{Learned Gate} directly predicts channel weights without solving the residual-fitting objective. \emph{Direct Output} predicts a bounded correction in velocity space, bypassing gated feature injection. \emph{Hidden-Only} removes the temporal video-state branch and predicts corrections from VLA hidden features. Their five-task means are 69.6\%, 78.8\%, and 79.6\%, respectively, versus 82.0\% for FAVR.

\subsection{State persistence versus static aggregation}
Window-wise resetting reinitializes the TTT state to learned $S_0$ before each new video window. It retains the same projections and within-window update mechanism but does not accumulate state across windows. Static pooling replaces recurrent writes with aggregate visual features. On the five-task experiment, these variants achieve 77.6\% and 68.8\%, compared with 82.0\% for full FAVR. This supports maintaining the recurrent TTT state across observation windows.

\subsection{Image-pretrained versus video-pretrained features}
On the four LIBERO suites, replacing V-JEPA with I-JEPA changes the mean from 98.6\% to 97.8\%. Table~\ref{tab:sup_representation} provides the individual suite results. The largest difference occurs on Long (+2.4 percentage points for V-JEPA); on Spatial, I-JEPA is higher by 0.4 points.

\begin{table}[t]
\centering\small
\caption{\textbf{Representation ablation on LIBERO: success rates (\%).} Both alternatives are used as frozen visual representations for the FAVR interface.}
\label{tab:sup_representation}
\setlength{\tabcolsep}{4pt}
\begin{tabular}{lrrrrr}
\toprule
Representation & Spatial & Object & Goal & Long & Avg.\\
\midrule
I-JEPA & 99.2 & 99.8 & 97.4 & 94.8 & 97.8\\
\textbf{V-JEPA} & 98.8 & 100.0 & 98.4 & 97.2 & \textbf{98.6}\\
\bottomrule
\end{tabular}
\end{table}

\subsection{Single-view versus three-view V-JEPA conditioning}
\label{app:multiview}
We study whether auxiliary video conditioning benefits from views that reveal local contact states. The frozen VLA policy itself receives three views in both configurations. Feeding only the head-camera history to V-JEPA gives 88.48\% mean success on the full 50-task RoboTwin benchmark, while head and both wrist-camera histories give 90.04\%. Both the head-only and three-view RoboTwin variants use a residual auxiliary loss weight of $0.5$.

The \texttt{open\_microwave} rollout illustrates one way the extra wrist evidence can matter. The wrist view reveals when the gripper slips from the handle, whereas a head view alone makes the contact change difficult to distinguish. A head-only visual correction may continue to favor nominal progression when regrasping is needed. The three-view configuration reaches 64\% success on this task versus 45\% for head-only, a 19-point difference. Synchronized head- and wrist-camera frames are shown in the main-paper figure.

\subsection{Joint and approximate gate outcomes}
The full FAVR solver retains the cross-channel interactions in $Q^\top Q$. FAVR-CF instead uses a diagonal gate followed by a regularized scalar rescaling (Sec.~\ref{app:favr_cf_proof}). In the main benchmark tables, FAVR-CF achieves 98.25\% on LIBERO and 84.55\% on LIBERO-Plus, compared with 98.60\% and 85.05\% for FAVR. Thus the approximate version trails joint optimization by 0.35 and 0.50 percentage points while remaining above the corresponding frozen baselines. Inference latency is reported in Sec.~\ref{app:favr_runtime}.

\section{Real-World Experimental Details}
\label{app:favr_realworld}

\subsection{Platform, data, and comparison protocol}
Real-world experiments use Cobot Magic, a dual-arm manipulation platform based on the Mobile ALOHA teleoperation framework~\cite{cobotmagic}. We collect 35 teleoperated demonstrations for each of four tasks. Two are standard manipulation tasks; two require information observed before the robot begins acting. The real-world visual-conditioning interface operates on the observed camera history while the underlying bimanual $\pi_{0.5}$ policy is frozen during FAVR adaptation. The platform records three RGB views: one fixed frontal external camera, a left wrist camera, and a right wrist camera, each at an original resolution of $640\times480$. The action executor publishes commands at $30\,$Hz. The policy predicts 50 action targets and executes the first ten before replanning. Each target is a 14-dimensional absolute-position command: six arm-joint targets (in radians) plus one continuous gripper-opening target for each arm. Demonstrations synchronously record all three camera streams, the actual follower-arm states, and the leader-arm teleoperation commands; supervised action labels are based on the leader commands, not obtained by copying follower states.

For standard manipulation, both comparison branches start from the \emph{same task-fine-tuned $\pi_{0.5}$ checkpoint}. One branch continues fine-tuning the base policy; the other freezes the base policy and V-JEPA and trains only the FAVR modules using the same number of additional optimizer steps. The reported success fractions are over 20 trials for each branch and task, as summarized in Table~\ref{tab:sup_realworld}.

\begin{table}[t]
\centering\small
\caption{\textbf{Standard real-world manipulation.} Each task has 35 demonstrations. Initial fine-tuning establishes the shared task-adapted checkpoint; continued fine-tuning and FAVR use the same additional update budget. Outcomes are successes out of 20 trials.}
\label{tab:sup_realworld}
\setlength{\tabcolsep}{4pt}
\resizebox{\columnwidth}{!}{%
\begin{tabular}{lrrrrr}
\toprule
Task & Initial & Extra & Initial FT & Cont. FT & FAVR \\
\midrule
Block pick-and-place & 5K & 3K & 10/20 & 12/20 & \textbf{15/20}\\
Paper-cup grasping & 15K & 5K & 8/20 & 11/20 & \textbf{13/20}\\
\bottomrule
\end{tabular}%
}
\end{table}

\subsection{Standard task definitions}
\textbf{Color-conditioned block pick-and-place.} The instruction identifies a target block by color. A successful episode requires selecting that block and placing it into the designated container. This task tests instruction-conditioned target selection, reaching, grasping, and placement in a coordinated sequence.

\textbf{Left/right paper-cup grasping.} The instruction specifies which of the two paper cups the robot should pick up. Compared with block pick-and-place, grasp success is sensitive to the end-effector approach and contact pose. The initial, continued-fine-tuned, and FAVR results in Table~\ref{tab:sup_realworld} follow the common 20-trial evaluation count for the respective tasks.

In the unadapted block policy, some failures involve choosing the incorrect color despite the pretrained vision--language backbone. The initial checkpoint records three such errors among its ten failures. 

\subsection{History-dependent tasks and observation/action separation}
\label{app:favr_history_tasks}
The history-dependent experiments are motivated by earlier uses of TTT as a recurrent memory mechanism, including its application to robotic interaction histories~\cite{ttt,robottt,tttrm}. We evaluate the history-conditioned selection enabled by the TTT temporal visual interface.

In \emph{covered-block retrieval}, one of two colored blocks is initially covered by a paper cup. After the observation phase, the cover is removed; the robot must identify and place the \emph{previously covered} block into a container. In \emph{history-conditioned cup selection}, the positions of two differently colored blocks are visible before separate cups cover them. After covering, the robot must choose the cup concealing the block named by the instruction. Both tasks are constructed so that the relevant identity or mapping is established during observation, not by the final visible arrangement alone.

The trial has two phases. During the \emph{observation phase}, a human manipulates the cups while the robot receives visual inputs and updates its TTT state but issues no movement commands. The \emph{action phase} begins only after the human withdraws from the workspace. Observation-stage motion is suppressed so that the robot cannot encode the target side in its proprioception through anticipatory arm movement.

Each history-dependent task uses 35 demonstrations and a 5,000-update FAVR adaptation. We report correct target selection. Table~\ref{tab:sup_history} gives the completed counts. Both tasks achieve 7/10 correct choices, above the 50\% binary chance level.

\begin{table}[t]
\centering\small
\caption{\textbf{History-conditioned target selection on the real robot.} Selection correctness is reported separately from full task completion.}
\label{tab:sup_history}
\resizebox{\columnwidth}{!}{%
\begin{tabular}{lccc}
\toprule
Task & Demos & Steps & Correct / trials \\
\midrule
Covered-block retrieval & 35 & 5K & 7/10\\
History-conditioned cup selection & 35 & 5K & 7/10\\
\bottomrule
\end{tabular}%
}
\end{table}

\subsection{Dependence on a task-capable initial policy}
A complementary control starts directly from the pretrained bimanual $\pi_{0.5}$ checkpoint, without first performing task-specific block-pick-and-place fine-tuning. Both the policy and video encoder remain frozen, and only the FAVR modules are trained for 5,000 updates on the block demonstrations. The resulting controller repeatedly moves its arms above the workspace without reliably approaching the designated block or establishing a stable grasp. The contrast with task-adapted initialization in Table~\ref{tab:sup_realworld} shows that FAVR improves an existing manipulation policy rather than learning base reaching and grasping skills from scratch.

\section{Inference Cost and Complexity}
\label{app:favr_runtime}

\subsection{Inference latency on RoboTwin 2.0}
Table~\ref{tab:sup_latency} reports the average inference latencies for the frozen policy, FAVR-CF, and FAVR. The two FAVR versions differ in the gate solver: FAVR employs joint FISTA iterations, while FAVR-CF uses a diagonal approximation followed by analytic scalar rescaling. The 16.6\,ms increase of FAVR-CF over the frozen policy corresponds to $16.6/1117.7=1.5\%$ (rounded); the $353.0$\,ms reduction relative to FAVR corresponds to $353.0/1487.3=23.7\%$ (rounded).

\begin{table}[t]
\centering\small
\caption{\textbf{Reported inference latency.} Measurements are averages over 100 inference calls under the pipelined schedule described below, using RoboTwin 2.0 inputs on an NVIDIA RTX PRO 6000 Blackwell GPU. Each call predicts a 50-action chunk with ten flow steps.}
\label{tab:sup_latency}
\setlength{\tabcolsep}{5pt}
\begin{tabular}{lr}
\toprule
Policy & Average latency (ms)\\
\midrule
Frozen VLA baseline & 1117.7\\
FAVR-CF & 1134.3\\
FAVR & 1487.3\\
\bottomrule
\end{tabular}
\end{table}

During action execution, the V-JEPA branch processes newly observed frames in ten-frame windows, with incremental TTT state updates. At replanning, processing of the remaining ten-frame window overlaps with VLA inference. The reported latencies are measured under this pipelined schedule.

\subsection{State storage and arithmetic}
Each TTT state contains $128^2$ FP32 scalars, or $128^2\times4=65{,}536$ bytes (64\,KiB). If committed and in-progress copies are kept simultaneously, the two matrices occupy 128\,KiB, excluding key/value activations, video frames, pretrained weights, and framework allocator overhead.

At fixed flow time, a joint gradient needs two products with the frozen $m\times d$ action-head matrix, which costs $O(md)$ arithmetic. With $N_{\mathrm{iter}}$ FISTA steps, the iterative gate therefore costs $O(N_{\mathrm{iter}}md)$ after computing its step-size information. The closed-form gate computes $s=x\odot(W^\top\mu)$ and $c^D=W(x\odot g^D)$ with cached column norms, costing $O(md)$ arithmetic and $O(d+m)$ extra working storage. These are gate-subroutine costs; they do not include V-JEPA encoding, TTT writes, backbone action inference, host synchronization, or physical robot execution.

\FloatBarrier
\clearpage
\section*{LLM Usage Statement}
OpenAI ChatGPT was used to assist with literature exploration, methodological discussions, manuscript drafting and revision, and checking equations and figures. The authors reviewed the assisted content and take full responsibility for the final manuscript and reported results.
{\small\bibliographystyle{ieeenat_fullname}\bibliography{references}}
\end{document}